\documentclass[journal]{IEEEtran}
\usepackage{hyperref}
\usepackage{amsmath}
\usepackage{tabularx}
\usepackage{graphicx}
\usepackage{amsthm}
\usepackage{amsthm}
 
\usepackage{algorithm2e}
\usepackage{amsthm}
\newtheorem{remark}{Remark}
\usepackage{bm}
\usepackage{xcolor}
\usepackage[short]{optidef}
\usepackage{amsfonts}
\usepackage{amssymb}
\usepackage{verbatim}
\usepackage{upgreek}
\usepackage{graphicx}
\usepackage{epstopdf}
\usepackage{subfigure}
\usepackage{color}
\usepackage{amsmath}
\usepackage{algorithmic}
\begin{document}

\title{An alternating peak-optimization method for optimal trajectory generation of quadrotor drones}

\author{Wytze A.B. de Vries, Ming Li, Qirui Song and Zhiyong Sun


\thanks{This work was supported in part by a starting grant from Eindhoven Artificial Intelligence Systems Institute (EAISI), Eindhoven, the Netherlands;
in part by EU-Horizon2020 - Marie Skłodowska-Curie Actions (MSCA SE)
grant No.101086228. 

The authors are with the Department of Electrical Engineering, Eindhoven University of Technology, and also with the Eindhoven Artificial Intelligence Systems Institute, PO Box 513, Eindhoven 5600 MB, The Netherlands. 
{\tt\small \{w.a.b.d.vries, q.song\}@student.tue.nl \{m.li3, z.sun\}@tue.nl}
}
}

\maketitle

\begin{abstract}
In this paper, we propose an alternating optimization method to address a time-optimal trajectory generation problem. Different from the existing solutions, our approach introduces a new formulation that minimizes the overall trajectory running time while maintaining the polynomial smoothness constraints and incorporating hard limits on motion derivatives to ensure feasibility. To address this problem, an alternating peak-optimization method is developed, which splits the optimization process into two sub-optimizations: the first sub-optimization optimizes polynomial coefficients for smoothness, and the second  sub-optimization adjusts the time allocated to each trajectory segment. These are alternated until a feasible minimum-time solution is found. We offer a comprehensive set of simulations and experiments to showcase the superior performance of our approach in comparison to existing methods. \\

A collection of demonstration videos with real drone flying experiments can be accessed at \url{https://www.youtube.com/playlist?list=PLQGtPFK17zUYkwFT-fr0a8E49R8Uq712l}. 
\\
\end{abstract}

\IEEEpeerreviewmaketitle

\section{Introduction}\label{Chapters:Introduction}

A quadrotor is a special type of unmanned aerial vehicle, which has received significant attention in recent years owing to its cost-effective design, ease of maintenance, and impressive maneuverability. Due to these advantages, quadrotors have found a wide range of applications, including environmental monitoring ~\cite{UAV_Environment}, agriculture ~\cite{UAV_Agriculture}, and search and rescue operations ~\cite{UAV_Rescue}. However, its operation with only four independent thrust forces leads to an underactuated system~\cite{Underactuated_Reasoning}. The inherent nature of the quadrotor, compounded by its complex nonlinear dynamics, strong coupling, and multi-variable actuation, presents a significant challenge when one seeks to optimize its motion for time efficiency~\cite{Foehn2021}.

Generally, time-optimal trajectory generation refers to the challenge of pushing a quadrotor to its theoretical limits, enabling the most aggressive motion possible. Within existing research, the continuous-time polynomials~\cite{Mellinger2011,Polynominal_Representation_1,Polynominal_Representation_2} serves as a prevalent approach for planning quadrotor trajectories. In particular, the trajectories are expressed as polynomial functions of the quadrotor's output variables, effectively leveraging the quadrotor's differential flatness property~\cite{Mellinger2011}. However, this approach has one major limitation, i.e., fixed running time, where the trajectory's overall running time is predetermined and not optimized for achieving the shortest possible duration, which means it is not time optimal. As an improvement, a novel time-optimal trajectory generation method is introduced in~\cite{Bry2015}. This method optimizes both the overall running time and trajectory polynomials, resulting in the generation of rapid and aggressive trajectories. However, this method fails to take control input limits into account, which is of great significance for many applications~\cite{Polynominal_Input_Constraints}. Furthermore, during the optimization process, it would stop when encountering a local minimum or reaching the limit of the maximum motor thrust. While this approach ensured that the drone utilized its maximum performance at a particular point in the trajectory, it did not guarantee the full utilization of available performance throughout the entire trajectory. 

\begin{figure}[tp]
    \centering
    \includegraphics[width=0.48\textwidth]{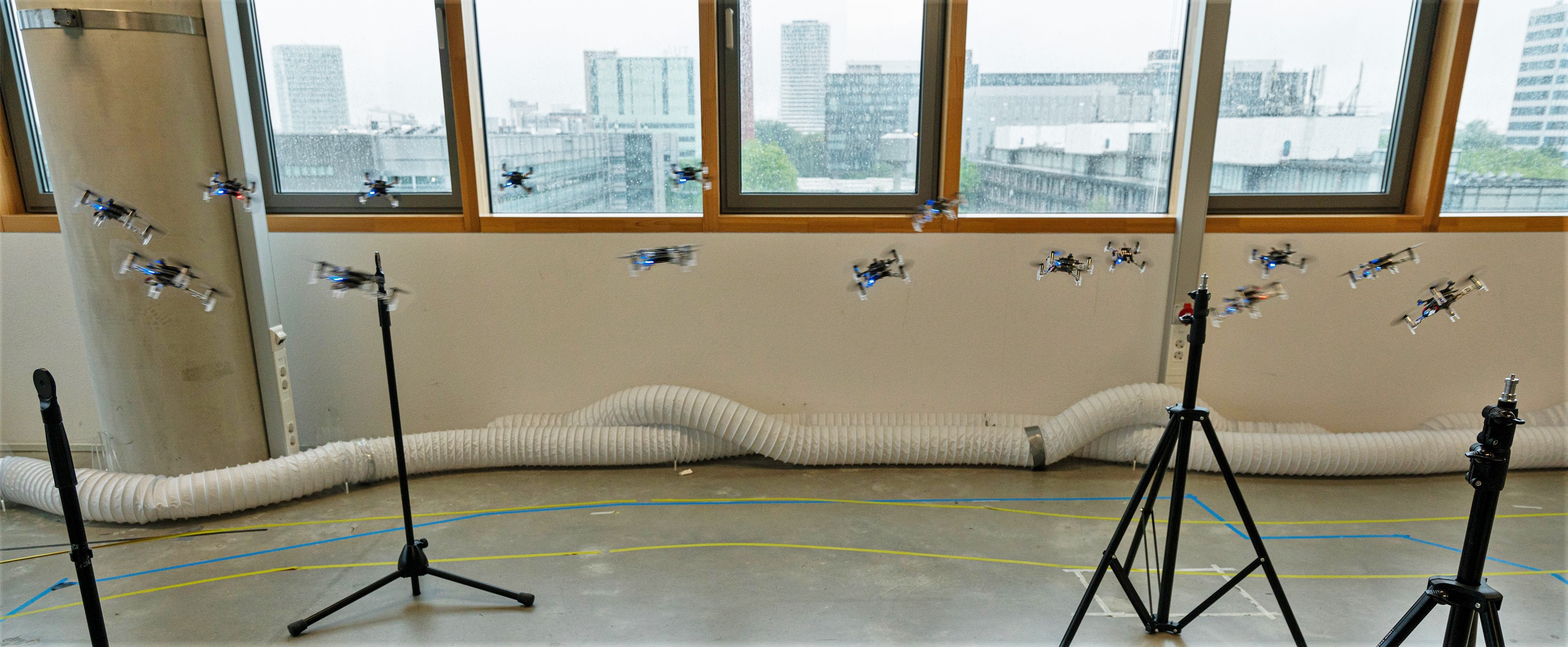}
    \caption{Composite image of a single drone completing a trajectory generated by the proposed peak-optimization method.}
    \label{fig:composite}
\end{figure}

In this paper, a novel time-optimal trajectory optimization is formulated, and a peak-optimization approach is proposed to address the time-optimal trajectory generation problem within the framework of continuous-time polynomials. A composite image of the drone flying in a physical world is exhibited in Fig.~\ref{fig:composite}, which showcases that our method can be implemented on an actual flying platform. Different from the formulation in~\cite{Bry2015}, our proposed optimization involves modifying the cost function to focus on minimizing the overall trajectory running time. While the polynomial constraints are retained to satisfy smoothness requirements, they are not optimized. Additionally, we explicitly set limits on the motion derivatives as hard constraints in the optimization. This guarantees that the generated time-optimal trajectories are feasible and can be executed by a given platform. To solve the optimization problem, an alternating peak-optimization method is developed. Specifically, the new optimization formulation is divided into two sub-optimization problems. For the first optimization problem, the overall running time is fixed and then the polynomial coefficients are deliberately selected to satisfy the smoothness constraints. For simplicity, we employ Mellinger's solution~\cite{Mellinger2011} directly to generate smooth polynomial trajectories in our implementation, as they inherently satisfy the required smoothness constraint. In the second optimization step, the polynomial coefficients remain fixed, while the time allocated to each segment of the trajectory is adjusted via a peak optimization technique. This adjustment process decreases the time intervals between segments, ensuring that the control input reaches its maximum limit. To demonstrate the advantages of our approach for generating time-optimal trajectories, our approach is compared with the existing trajectory generation methods through numerous simulation and experiment results via a platform called Crazyflie~\cite{Crazyflie}. 


The structure of this paper is as follows. In Section~\ref{Chapters:DiffFlat}, we introduce the pertinent drone dynamics, explore the concept of differential flatness, and provide a framework for continuous-time polynomial trajectory optimization. The proposed new method to optimize the segment times is presented in Section \ref{Chapters:New}. Comparisons and performance of time-optimal  trajectories generated by different methods are evaluated in Section \ref{Chapters:SimulationComparison}. Experimental results of the proposed methods are presented in Section \ref{Chapters:Results} and the paper is concluded in Section~\ref{Chapters:Discussion}.


\section{Preliminary on trajectory optimization}\label{Chapters:DiffFlat}
\subsection{Quadrotor dynamics and differential flatness}
\subsubsection{Quadrotor dynamics}
The quadrotor is modeled as a rigid body with four rotors.  The motion of the quadrotor is described by the following equations~\cite{Quadrotor_Model}: 
\begin{equation}\label{Quadrotor_model}
\begin{split}
\dot{\mathbf{r}} &=\mathbf{v},\qquad\qquad\quad\,\,\,\, m \dot{\mathbf{v}} =-m g \mathbf{e}_{3}+f \mathbf{R} \mathbf{e}_{3}, \\
\dot{\mathbf{R}}&=\mathbf{R}\bm{\Omega}^{\times},\qquad\qquad\,\,
\mathbf{J}\dot{\bm{\Omega}}=-\bm{\Omega}\times\mathbf{J}\bm{\Omega}+\bm{\tau},
\end{split}
\end{equation}
where the position and velocity of the quadrotor are denoted by $\mathbf{r}=[x,y,z]^{\top}\in\mathbb{R}^{3}$ and $\mathbf{v}=[v_x,v_y,v_z]^{\top}\in\mathbb{R}^{3}$, respectively. The rotation matrix $\mathbf{R}$   for a quadrotor is obtained using Euler angles $\bm{\zeta}=[ \phi, \theta, \psi]^{\top}$, which represent three elemental rotations about the body-fixed axes $z, x, y$. In the body frame, the angular velocity is denoted by $\bm{\Omega}=[p,q,r]^{\top}$, and the skew-symmetric matrix $\bm{\Omega}^{\times}$ is defined as follows: for any vector $\mathbf{s}\in\mathbb{R}^{3}$, $\bm{\Omega}^{\times}\mathbf{s}=\bm{\Omega}\times\mathbf{s}$, where $\times$ denotes the vector cross product. We thereby define the state $\mathbf{x}=[\mathbf{r}^{\top},\mathbf{v}^{\top},\bm{\zeta}^{\top},\bm{\Omega}]^{\top}$. The mass of the quadrotor is denoted by $m$. The unit vector $\mathbf{e}_{3}=[0,0,1]^{\top}$ specifies the direction of the gravitational force $g$ in the inertial frame $\mathcal{F}_{\mathcal{W}}$. The quadrotor's total force, represented by $f\in\mathbb{R}^{+}$, is the combined effect of the individual thrust forces produced by its rotors. The total torque $\bm{\tau}$, which is generated by the quadrotor's rotors, determines its rotational motion and orientation. The quadrotor's inertia matrix, represented by $\mathbf{J}=\mathrm{diag}\left([\mathrm{J}_{x}, \mathrm{J}_{y}, \mathrm{J}_{z}]\right)$, describes how the mass is distributed throughout the quadrotor and determines how it responds to external forces and torques.
\subsubsection{Differential flatness}
For a quadrotor, its flat output is defined as follows \cite{Mellinger2011} 
\begin{equation}
    \bm{\sigma} = 
    \begin{bmatrix}
    \textbf{r} , \psi \\
    \end{bmatrix}^\top = 
    \begin{bmatrix}
    x , y , z , \psi \\
    \end{bmatrix}^\top.
\end{equation}
Differential flatness shows that the inputs $\textbf{u}$ and states $\textbf{x}$ of the drone are algebraically linked to its trajectory, described by four flat outputs $\sigma$ and its derivatives.
\begin{equation}
    \begin{split}
    \mathbf{x}&=\varphi_{\mathbf{x}}(\bm{\sigma},\dot{\bm{\sigma}},\cdots,\bm{\sigma}^{(s-1)}),\\
    \mathbf{u}&=\varphi_{\mathbf{u}}(\bm{\sigma},\dot{\bm{\sigma}},\cdots,\bm{\sigma}^{(s-1)}),
    \end{split}
\end{equation}
where $s$ indicates the $s$-th time derivative of a function, $\varphi_{\mathbf{x}}: (\mathbb{R}^{m})^{s-1}\mapsto\mathbb{R}^{n}$ and $\varphi_{\mathbf{x}}: (\mathbb{R}^{m})^{s}\mapsto\mathbb{R}^{m}$ are both induced by the dynamics of the quadrotor.  
\subsection{Formulation of trajectory optimization}
\subsubsection{Polynomial trajectory optimization}
When assessing the evaluation of a flat output variable denoted as $\bm{\sigma}$ within a time interval $t=[0,\tau]$ defined by a polynomial $P(t)$ connecting two points in the flat output space, each flat output and segment trajectory are represented by a polynomial $P(t)=\sum\limits_{i=0}^{N}{p}_{i}t^{i}$. The optimization of the coefficients for this polynomial can be achieved by minimizing a cost function $J$ given by: 
\begin{equation}
    J = \int_{t=0}^{t=\tau} c_\mathrm{0} P(t)^2 + c_\mathrm{1} \dot{P}(t)^2 + c_\mathrm{2} \ddot{P}(t)^2 + ... + c_N P^{(N)} (t)^2 dt,  
    \label{eq:minfuncP}
\end{equation}
where $c_i, i=0,\cdots, N$ denotes the weight of each derivative. This allows for the minimization of the total speed, acceleration, jerk, or snap present during the trajectory. 
This cost function can be rewritten into a quadratic form as 
\begin{equation}
    J^k = \mathbf{p}_k^\top \mathbf{Q}_k \mathbf{p}_k,
\end{equation}
where $\textbf{p}_\textbf{k}$ is a vector which contains the coefficients of the polynomial and $\textbf{Q}_\textbf{k}$ is the cost matrix. The computation of the cost matrix in the quadratic form was explained in \cite{Bry2015}. Since trajectories often consist of multiple segments, a more comprehensive cost function is necessary to encompass all coefficients and cost matrices. This extended cost function can be succinctly expressed as:
\begin{equation}
    J = 
    \begin{bmatrix}
    \mathbf{p}_1 \\
    \vdots \\
    \mathbf{p}_K
    \end{bmatrix}^\top
    \begin{bmatrix}
    \textbf{Q}_1 & & \\
    & \ddots & \\
    & & \textbf{Q}_K \\
    \end{bmatrix}
        \begin{bmatrix}
    \mathbf{p}_1 \\
    \vdots \\
   \mathbf{p}_K
    \end{bmatrix},
    \label{eq:minfunc}
\end{equation}
where K denotes the number of the polynomial segments. By concatenating the desired constraints on the initial and final derivatives of the polynomial, we have a standard quadratic programming (QP) problem

\begin{mini!}
{\mathbf{p}_k}
{J} {\label{eq:mellinger_poly}}{}
\addConstraint{\mathbf{A}\mathbf{p}-\mathbf{b}}{ = 0}
\end{mini!}
where $\mathbf{p}=[\mathbf{p}_{1}^{\top},\cdots\mathbf{p}_{K}^{\top}]^{\top}$, $\textbf{A}$ and $\textbf{b}$ are used to constrain the start, end or any intermediate positions, velocities or other derivatives and is also used to ensure continuity of any derivative. 
\subsubsection{Optimal trajectory with alternating optimization}
The previously discussed method does not guarantee the minimal possible cost for the given problem since the segment times were initially fixed to construct the cost function. A solution with lower costs could probably be found by redistributing the time among the segments differently. Hence, it becomes essential to optimize both the time distribution and polynomial coefficients to achieve the most cost-effective solution. Mellinger and Kumar \cite{Mellinger2011} (referred to as the ``Mellinger method" hereafter) proposed an alternating optimization approach to address this challenge. This approach employs a gradient descent method, starting with an initial estimate of the segment times and iteratively optimizing the polynomial coefficients. 
\begin{mini!}
{t_\mathrm{k}}
{f(\mathbf{t})} {\label{eq:timecost}}{}
\addConstraint{\sum_{\mathrm{k}=1}^K t_\mathrm{k}}{ = T}
\addConstraint{t_\mathrm{k}}{\geq 0},
\end{mini!}
where $f(\cdot)$ is the cost function with the optimised coefficients, $\mathbf{t} = [t_1, \cdots, t_\mathrm{K}]^\top$, and $T$ is the total time. 
\subsubsection{Optimal trajectory with aggressive motion}
An improved method to obtain aggressive motion was proposed by Bry et al. \cite{Bry2015} (hereon referred to as ``Bry method"). It removes the total time constrained in Eq.~\eqref{eq:timecost} so that the total time would decrease. Then the optimization function was reformulated so that both the polynomial coefficients and the segment times are optimization variables. The cost function was expanded and a cost was placed upon the total time, which was described by
\begin{equation}
    J = 
    \begin{bmatrix}
    \mathbf{p}_1 \\
    \vdots \\
    \mathbf{p}_K
    \end{bmatrix}^\top
    \begin{bmatrix}
    \mathbf{Q}_1(\tau_1) & & \\
    & \ddots & \\
    & & \textbf{Q}_K(\tau_K) \\
    \end{bmatrix}
        \begin{bmatrix}
    \mathbf{p}_1 \\
    \vdots \\
   \mathbf{p}_K
    \end{bmatrix}
    + c_t \sum_{\mathrm{k}=1}^K \tau_\mathrm{k},
    \label{eq:BryOpt}
\end{equation}
where $c_t$ is a weighting coefficient on the total time. This would lead to an increase in total jerk or snap, but a lower total time, which results in more aggressive and faster trajectories. The weighting coefficient could be increased if a more aggressive trajectory was desired and lowered if a trajectory with less total snap was desired. 

It should be noted that the most aggressive and feasible trajectory is inherently non-smooth and would therefore not be achievable using continuous-time polynomial trajectories. Some other form of optimization such as time-optimal planning is required to achieve the most aggressive trajectory possible as was demonstrated by Foehn, Romero, and Scaramuzza in \cite{Foehn2021}. This approach does however require a very complex and accurate model of the drone, while the continuous-time polynomial trajectories require far less information and are therefore easier to implement.


\section{A novel trajectory optimization formulation and the peak-optimization method}\label{Chapters:New}
In this section, we introduce a new time-optimal trajectory formulation that prioritizes minimizing the overall trajectory running time while preserving polynomial smoothness constraints. We also establish limits on the motion derivatives as hard constraints, which ensure the feasibility of generated time-optimal trajectories. To improve the efficiency of our optimization approach, we propose the ``peak optimization" method, which splits the problem into two sub-optimization steps. Firstly, a continuously smooth trajectory with a fixed time duration is created, followed by a an optimization that optimizes the segment times using a gradient descent approach. Then the two sub-optimizations are alternated until a feasible solution is found that achieves the minimal total running time. 

\subsection{A novel formulation of time-optimal trajectory optimization}
 We define the most aggressive trajectory possible as the trajectory that reaches all waypoints in the shortest time while staying within the limitations of the drone. There are multiple limitations that the feasible trajectories have to respect, such as maximum thrust, maximum power, and aerodynamic drag. These limitations can be mathematically expressed in motion derivatives by using the property of differential flatness. Identifying these exact limits is a very challenging problem as it depends on many parameters. The thrust that the motor is able to produce depends on the command signal and the battery voltage which changes during flight \cite{Shi2022}. The thrust and torque are coupled \cite{Foehn2021} which means that the acceleration limits and snap limits are coupled as well. 
 Our goal is to obtain a trajectory that is both time-optimal, smooth and feasible. Therefore a new optimization formulation is presented as follows
\begin{mini}
{t_\mathrm{k}, \textbf{p}_\mathrm{k}}
{\sum_{\mathrm{k}=1}^{\mathrm{K}} t_\mathrm{k}} {}{\label{eq:peak_opt_complete}}
\addConstraint{\rho_\mathrm{s}-\| \frac{\partial^\mathrm{s}\textbf{r}_\mathrm{k}(t)}{\partial t^\mathrm{s}} \|}{  \geq 0 ~\forall \mathrm{s} }
\addConstraint{\textbf{Ap} - \textbf{b}}{ = 0},
\end{mini}
where $t_\mathrm{k}$ is the time duration of segment $\mathrm{k} = [1,\cdots, K]$, $s$ is the order of the motion derivative and $\rho_\mathrm{s}$ is the limit of the $s$th motion derivative. Here, the cost is placed only on the total time, but the constraints ensure feasibility and that the polynomial trajectory is continuous and intersects all waypoints.

The solution to this problem would require that the time and polynomial coefficients be optimized simultaneously. This is however a very difficult problem to solve analytically as it is non-linear and non-convex. Therefore we adopt the alternating approach as Mellinger's method as discussed in Section \ref{Chapters:DiffFlat}, splitting the optimization into two sub-optimization problems involving the iterative optimization of polynomial coefficients $\mathbf{p}_\mathrm{k}$ and segment times $t_\mathrm{k}$. Specifically, we introduce the alternating peak-optimization method in the next subsection.

\subsection{Alternating peak-optimization approach}
\begin{figure}[t]
\hspace*{-0.5cm}
    \centering
    \includegraphics[width=0.5\textwidth]{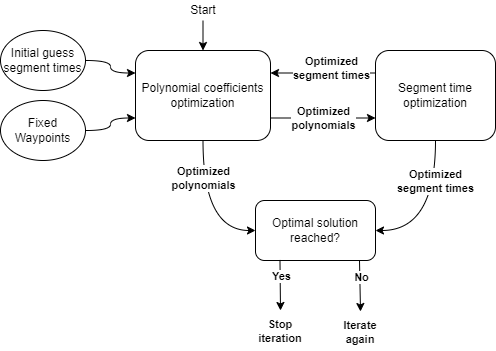}
    \caption{Schematic overview of the peak optimization method}
    \label{fig:Schematic}
\end{figure}

Generally, the alternating peak-optimization approach involves splitting the optimization problem in Eq.~(\ref{eq:peak_opt_complete}) into two distinct optimization processes. A schematic overview of the proposed approach is shown in Fig. ~\ref{fig:Schematic}.

\subsubsection{First sub-optimization problem} 
 During the first optimization, the segment times are fixed and then the polynomial coefficients are selected to satisfy the second constraint of Eq. (~\ref{eq:peak_opt_complete}). Mellingers optimization as described in Eq. (~\ref{eq:mellinger_poly}) is used to generate a new smooth polynomial trajectory, as it inherintly satisfies the smoothness constraints.

 \subsubsection{Second sub-optimization problem} 
During the second optimization, the polynomials are fixed and the segment times are then updated by solving the following problem using gradient descent
\begin{mini}
{t_\mathrm{k}}
{\sum_{\mathrm{k}=1}^{\mathrm{K}} t_\mathrm{k}} {}{\label{eq:opt_time}}
\addConstraint{\rho_\mathrm{s}-\| \frac{\partial^\mathrm{s}\textbf{r}_\mathrm{k}(t)}{\partial t^\mathrm{s}} \|}{  \geq 0 ~\forall \mathrm{s},}
\end{mini}

This is the same problem as Eq.~(\ref{eq:peak_opt_complete}), but with the second constraint removed that was already satisfied by the polynomial coefficient optimization. The segment times can either be decreased to minimise the cost function of Eq.~(\ref{eq:peak_opt_complete}), or be increased to ensure feasibility by satisfying the first constraint of Eq.~(\ref{eq:peak_opt_complete}).

The complexity of checking the constraint of Eq.~(\ref{eq:opt_time}) can be significantly decreased by reducing the time dependent norm of the motion derivative to a single scalar per segment by first computing the peak value of each motion derivative during each segment. This peak $\kappa_\mathrm{k}$ is normalised with the limit $\rho_\mathrm{s}$ and calculated for each segment $k$ by

\begin{equation}
    \kappa_\mathrm{k}(\mathrm{s}) = \frac{max(\| \frac{\partial^\mathrm{s}\textbf{r}_\mathrm{k}(t)}{\partial t^\mathrm{s}} \|)}{\rho_\mathrm{s}},
\label{eq:kapp_i}
\end{equation}
where $s = [1,2,3,4]$ as the motion derivatives up to the fourth order are considered. Therefore $\kappa > 1$  when its limit is exceeded and $\kappa < 1$ when it is below its limit. These peak values can then be used to update the current segment times $t_\mathrm{k,old}$ to the new segment times $t_\mathrm{k,new}$ so that the trajectory is as close to its most restrictive limit as possible using

\begin{equation}
    t_\mathrm{k,new} = t_\mathrm{k,old}\cdot(1-\delta_{\mathrm{1}}\cdot(1-max(\bm{\kappa}_\mathrm{k}))),
\label{eq:t_i_new}
\end{equation}
where $\bm{\kappa}_\mathrm{k} = [\kappa_\mathrm{k}(1), \cdots,  \kappa_\mathrm{k}(4)]^\top$ and $\delta_{\mathrm{1}}$ is the first learning rate. The learning rate was added as the shape of the trajectory (and with it the shape the derivatives) will change in a non-linearly when the segment times are adjusted. Therefore the segment times are adjusted in smaller steps during each iteration. The learning rate can be increased to improve computational efficiency. However, as the motion derivatives do not scale linearly with the change in segment times, it is nearly impossible to arrive at the exact limit for each segment using this iterative method due to the continuity constraints of Eq.~(\ref{eq:mellinger_poly}). The method is therefore iterated until the largest value in $\bm{\kappa}_\mathrm{k}$ are within the range of $[0.95 ~ 1.05]$ for each segment. Increasing the learning rate too much can lead to the solution never meeting this criteria due to it always over correcting. Therefore a trade-off has to be found for the learning rate between efficiency and a guarantee that the optimization will converge to a final solution.


\subsection{Discussions on the alternating optimization and implementation}
\begin{figure}[t]
\hspace*{-0.5cm}
    \centering
    \includegraphics[width=0.5\textwidth]{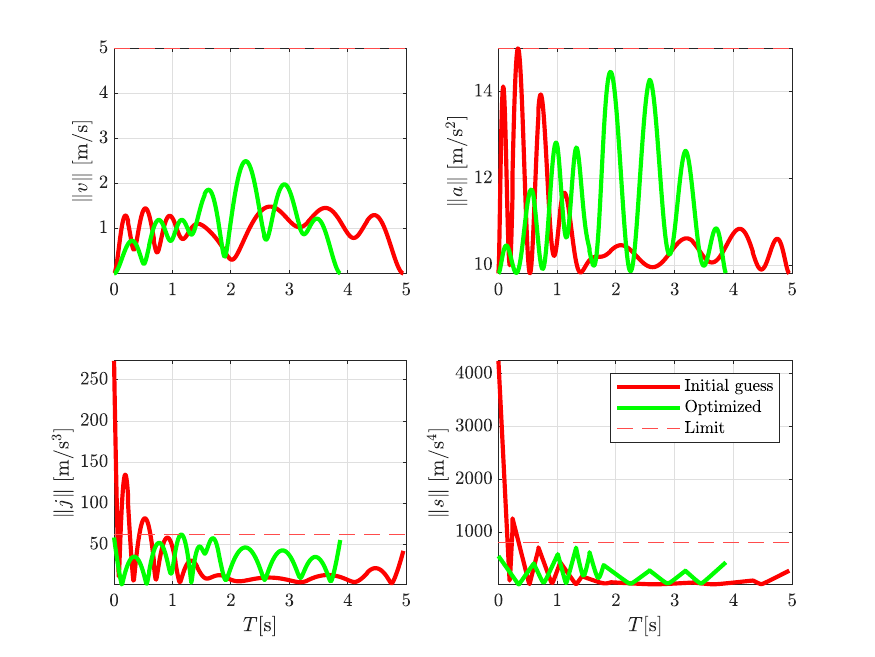}
    \caption{Motion derivatives of an example trajectory}
    \label{fig:Opt_derivatives}
\end{figure}

\begin{figure}[t]
\hspace*{-0.5cm}
    \centering
    \includegraphics[width=0.5\textwidth]{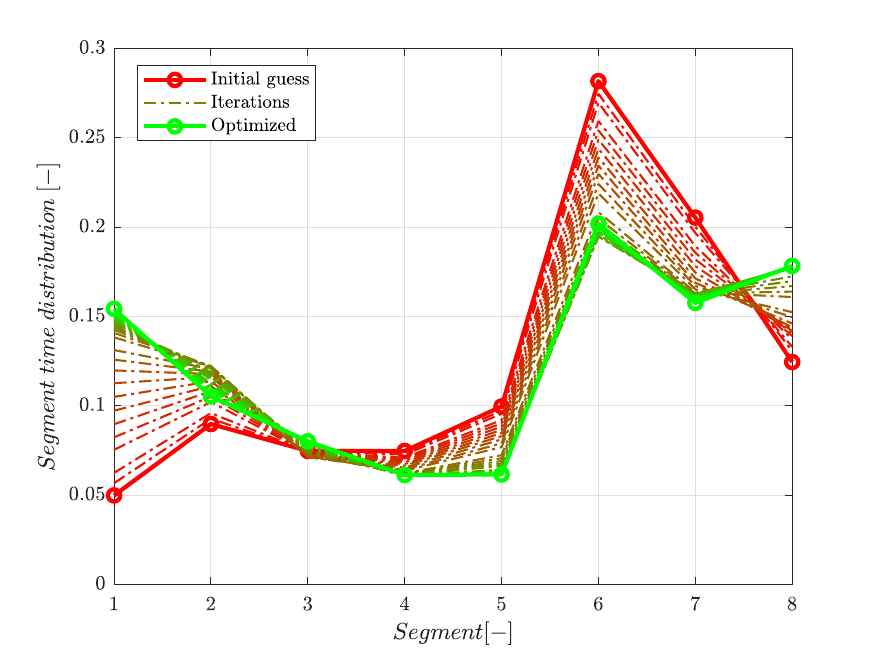}
    \caption{Iterative optimized segment times of an example trajectory}
    \label{fig:Opt_SegTimes}
\end{figure}

\begin{figure}[t]
    \centering
    \includegraphics[width=0.5\textwidth]{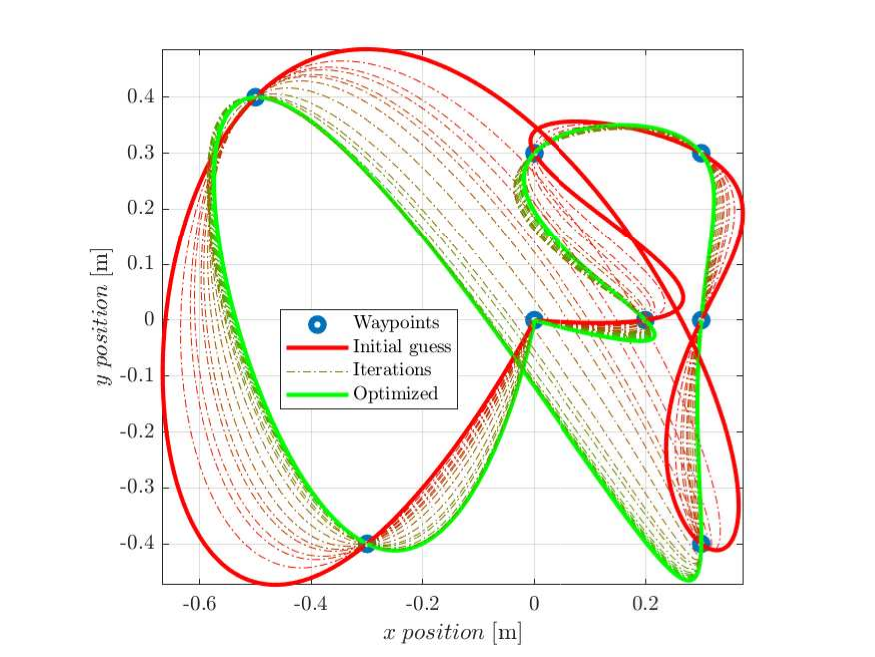}
    \caption{Iterative optimized example trajectory}
    \label{fig:Opt_Traj}
\end{figure}

The proposed method is illustrated using Fig.~\ref{fig:Opt_derivatives}. Here, the velocity, acceleration, jerk and snap of a continuous-time polynomial trajectory that is optimised for minimum jerk are shown that have been generated using an initial guess for the segment times. The initial guess for the segment times is proportional to the euclidean distance between the waypoints and the total time was set to 5 seconds. This set of waypoints was chosen for this example, because it has to manoeuvre a tight section at the start that will show large values for jerk and snap, while the rest of the trajectory requires a lot of acceleration, but not a lot of jerk and snap due to the long straight sections and gentle curves. The limits on jerk and snap are exceeded during the first second, which will make the trajectory infeasible. It is however far below any limit during the latter 4 seconds of this trajectory. 
The figure shows that the optimised trajectory is feasible and over a second faster than the infeasible initial guess. Figs.~\ref{fig:Opt_SegTimes} and~\ref{fig:Opt_Traj} show that it is done by allocating more time to the tight and twisty first two segments while reducing the segment times for the latter segments. Therefore the large peaks in jerk and snap are removed for the first segments, the acceleration and speeds are increased for the latter segments and the trajectory is significantly shortened as well.

\begin{remark}
The acceleration is limited by the thrust the drone is able to produce. Due to gravity, the drone is able to climb a lot slower than it is able to drop. Therefore $\kappa_\mathrm{k}(2)$ is adjusted to make it represent the mass normalised force that the drone has to produce to execute the trajectory. This is therefore represented by
\begin{equation}
    \kappa_\mathrm{k}(2) = \frac{max(\sqrt{(\frac{\partial^2x_\mathrm{k}}{\partial t^2})^2+(\frac{\partial^2y_\mathrm{k}}{\partial t^2})^2+(\frac{\partial^2z_\mathrm{k}}{\partial t^2}+g)^2})}{\rho_\mathrm{s}},
\label{eq:kappa2}
\end{equation}
where $g$ is the gravitational constant. 
\end{remark}
\begin{remark}
The continuity constraints make the solution of a segment depend on its neighbouring segments. For example, if a sharp turn has to be made during a segment it will require much higher accelerations when it enters the segment with a high velocity compared to when it enters with a low velocity. Therefore it would beneficial to not only increase the segment time during the current segment, but also its neighbouring segments. This is done according to

\begin{equation}
    t_\mathrm{k\pm1,new} = t_\mathrm{k\pm1,old}\cdot(1-\delta_{\mathrm{2}}\cdot(1-max(\bm{\kappa}_\mathrm{k}))),
\label{eq:t_i_new_pm}
\end{equation}
where $\delta_{\mathrm{2}}$ is the second learning rate.
\end{remark}

\begin{remark}
Increasing the segment time does not guarantee a decrease in all of the motion derivatives. For example, an increase in time often leads to a larger arc when the polynomial optimization is optimised for snap, which does not lower the accelerations. This can be solved by scaling all the segment times uniformly when any of the limits are exceeded. This preserves the shape of the trajectory while lowering all the derivatives. This is done according to
\begin{equation}
    \textbf{t}_\mathrm{new} = \textbf{t}_\mathrm{old}\cdot(1-\delta_{\mathrm{1}}\cdot(1-max(\bm{\kappa}))),
\label{eq:t_i_new_all}
\end{equation}
where $\textbf{t}_\mathrm{new}$ and $\textbf{t}_\mathrm{old}$ are vectors of all the new and old segment times respectively, and $\bm{\kappa} = [\bm{\kappa}_\mathrm{1}, \cdots, \bm{\kappa}_\mathrm{n_{segments}}]$ is a matrix of all the limit normalised peak values for all motion derivatives and all segments. Then the time optimization can continue as described earlier. 
\end{remark}
An overview of the segment time optimisation is shown in Alg.~\ref{alg:peaktime}. The different methods will be evaluated further in the next section.

\RestyleAlgo{ruled}
\begin{algorithm}
\caption{Segment time optimization}\label{alg:peaktime}
\KwData{$\textbf{t}_\mathrm{old}, \textbf{p}(t)$, $\rho$, }
\KwResult{$\textbf{t}_\mathrm{new}$}

\For{$\mathrm{k}\leftarrow 1$ \KwTo $\mathrm{K}$}{
    \For{$\mathrm{s}\leftarrow 1$ \KwTo $4$}{
        \eIf{$\mathrm{s}==2$}{
        $\kappa_\mathrm{k}(\mathrm{s}) \gets Eq.~(\ref{eq:kappa2})$\; 
        }{
        $\kappa_\mathrm{k}(\mathrm{s}) \gets Eq.~(\ref{eq:kapp_i})$\;
        }
    }   
}

\eIf{$\kappa_\mathrm{k} \geq 1$}{
    $\textbf{t}_\mathrm{new} \gets Eq.~(\ref{eq:t_i_new_all})$\; 
    }{
    \For{$\mathrm{k}\leftarrow 1$ \KwTo $\mathrm{K}$}{
        $t_\mathrm{k,new} \gets Eq.~(\ref{eq:t_i_new})$\;
        $t_\mathrm{k\pm1,new} \gets Eq.~(\ref{eq:t_i_new_pm})$\;    
    }
}
\end{algorithm}

\section{Comparison in simulation}\label{Chapters:SimulationComparison}
The peak optimization method was implemented and compared to Mellingers method and the initial guess. The waypoints as described in the previous sections were used again as it shows the challenge of being limited by different motion derivatives during the duration of the trajectory. All trajectories start with the same initial guess of segment times as described in the previous section. The trajectories shown were generated to achieve the lowest total running time and were validated to be feasible during experiments which will be discussed in the next section. All trajectories used fifth order polynomials. 

\begin{figure}[t]
    \centering
    \includegraphics[width = 0.5\textwidth]{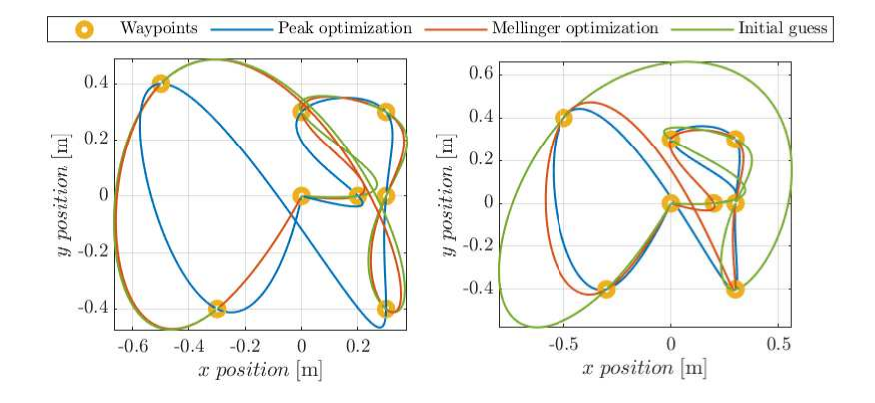}
    \caption{Comparison of minimum jerk (left) and minimum snap (right) trajectories}
    \label{fig:TrajSim}
\end{figure}

\begin{figure}[t] 
    \centering
    \includegraphics[width=0.5\textwidth]{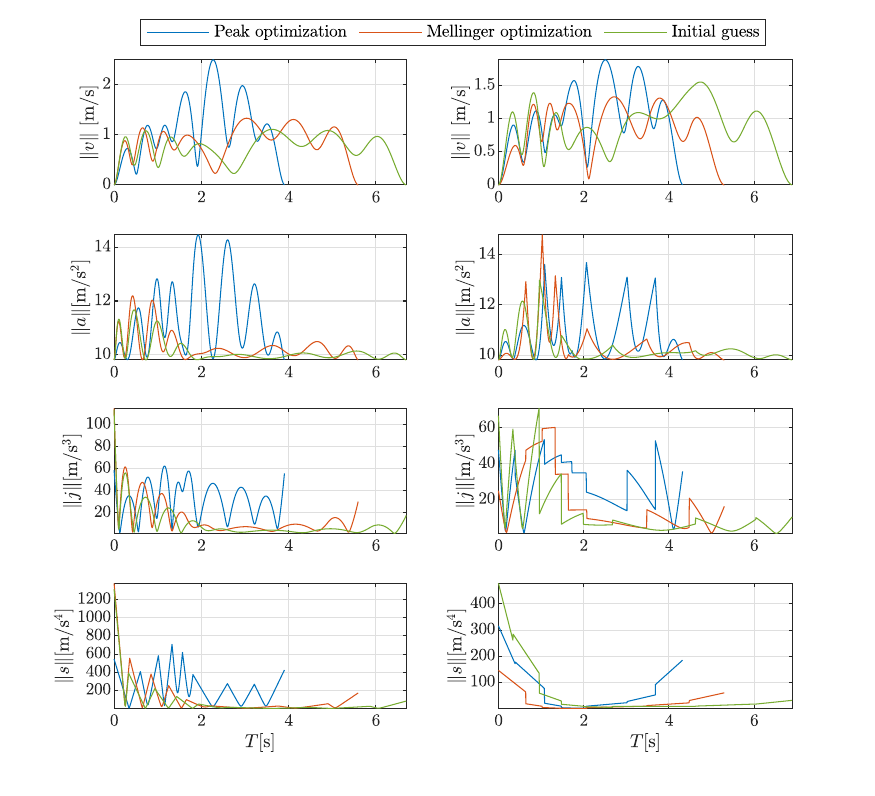}
    \caption{Motion derivatives of minimum jerk (left) and minimum snap (right) trajectories.}
    \label{fig:DerSim}
\end{figure}

\begin{figure}[t]
    \centering
    \includegraphics[width = 0.45\textwidth]{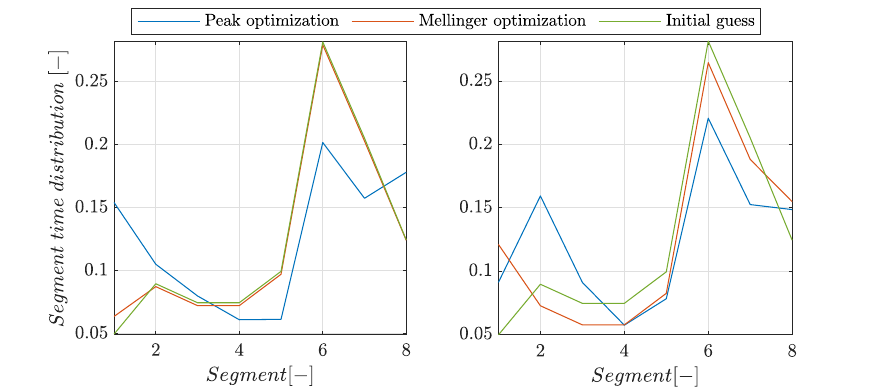}
    \caption{Segment time distribution of minimum jerk (left) and minimum snap (right) trajectories}
    \label{fig:SegTimeSim}
\end{figure}

The resulting trajectories can be seen in Fig.~\ref{fig:TrajSim}, the motion derivatives are shown in Fig.~\ref{fig:DerSim}, and the segment time distribution is shown in Fig.~\ref{fig:SegTimeSim}. 

\subsection{Minimum jerk trajectories}
The peak optimization was generated using the limits of 5 $m/s$, 15.5 $m/s^2$, 62 $m/s^3$ and 800 $m/s^4$ and resulted in a total time of 3.9 seconds. The Mellinger optimization used a total time of 5.6 seconds, and the initial guess used a total time of 6.7 seconds. The three different optimization methods show similar magnitudes of speed, acceleration, jerk, and snap during the first second. However these magnitudes decrease significantly after that for the initial guess and the Mellinger optimization, while the magnitudes for the peak optimization are significantly higher. The reason for this can explained with the segment time distribution as the peak optimization allocates significantly less time to the latter segments and more to the first three segments. Here it can also be seen that the time segment distribution from the Mellinger optimization is not very different from the initial guess as it settled quickly at a local minima. It should be noted that the Mellinger optimization is reliant on a good initial guess as it does not give guarantee of global optimality due to its gradient descent based approach. The maximum speed reached with peak optimization is 125\% and 88\% higher than the initial guess and the Mellinger optimization respectively. That, combined with the shorter path taken through the waypoints, result in a reduction of the total by 42\% and 30\% compared to the initial guess and the Mellinger optimization respectively.

\subsection{Minimum snap trajectories}
The results are similar for the minimum snap trajectories. Here, the peak optimization was generated using the limits of 5 $m/s$, 14.5 $m/s^2$, 54 $m/s^3$ and 800 $m/s^4$ and resulted in a total time of 4.3 seconds. The Mellinger optimization used a total time of 5.3 seconds, and the initial guess used a total time of 6.9 seconds. 
It can be seen that the peak optimised trajectory actually does not meet these limits consistently. This is due to the last two remarks discussed in the previous section. The optimization was stopped before the limit was reached for each segment, as it was not able to find a feasible trajectory with a shorter total time.
It is however still able to produce a significantly faster trajectory than the other two methods. Again, by allocating significantly more time to the first three segments and less to the latter segments. 
The Mellinger optimization goes in an opposite direction during the second and third segment by allocating less time than the initial guess compared to the peak optimization. However, both methods generate a shorter total time than the initial guess by 38\% and 23\% for the peak optimization and Mellinger optimization respectively.



\section{Implementation and results}\label{Chapters:Results}
The trajectories shown in the previous section were flown using the crazyswarm \cite{Preiss2017} package combined with the VICON motion capture system and the Crazyflie 2.1 drones \cite{Giernacki2017}. 

\begin{figure}[t]
    \centering
    \includegraphics[width = 0.5\textwidth]{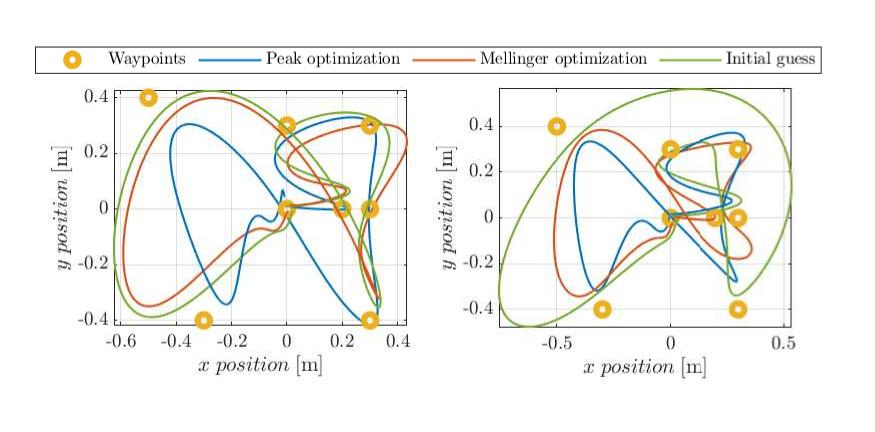}
    \caption{Experimental results, comparison of minimum jerk (left) and minimum snap (right) trajectories}
    \label{fig:TrajRes}
\end{figure}

\begin{figure}[t] 
    \centering
    \includegraphics[width=0.5\textwidth]{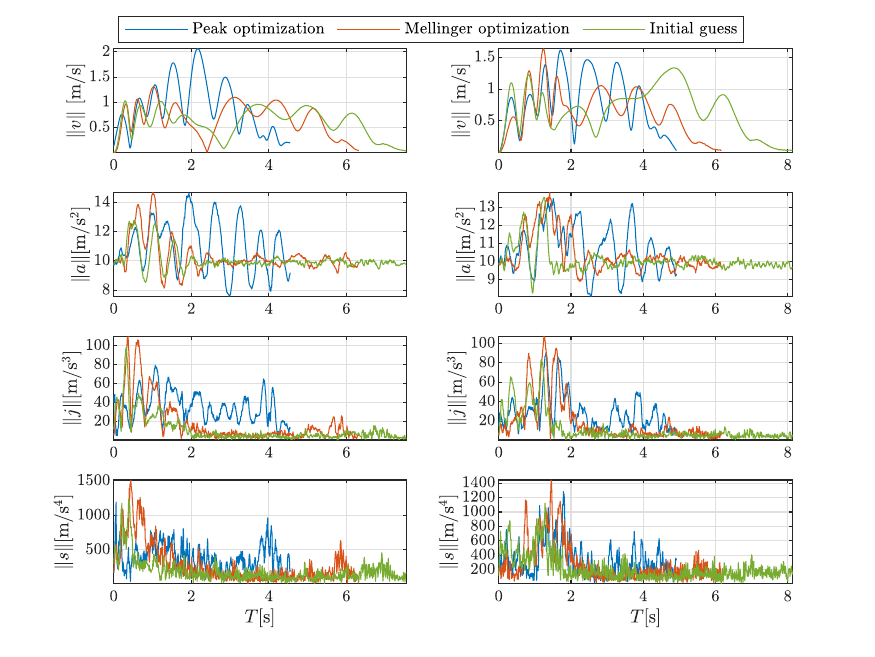}
    \caption{Experimental results, motion derivatives of minimum jerk (left) and minimum snap (right) trajectories.}
    \label{fig:DerRes}
\end{figure}

The data was collected using the VICON system and the flown trajectories are shown in Fig. ~\ref{fig:TrajRes}, while the motion derivatives are shown in Fig. ~\ref{fig:DerRes} and the resulting total times are summarised in Table ~\ref{tab:finaltime}. The motion derivatives are calculated by differentiating the position recorded by the VICON system, which was sampled at 100Hz, and smoothed using a 5 point averaging filter for each differentiation step. 

\subsection{Minimum jerk trajectories}
\begin{figure}[t] 
    \centering
    \includegraphics[width=0.5\textwidth]{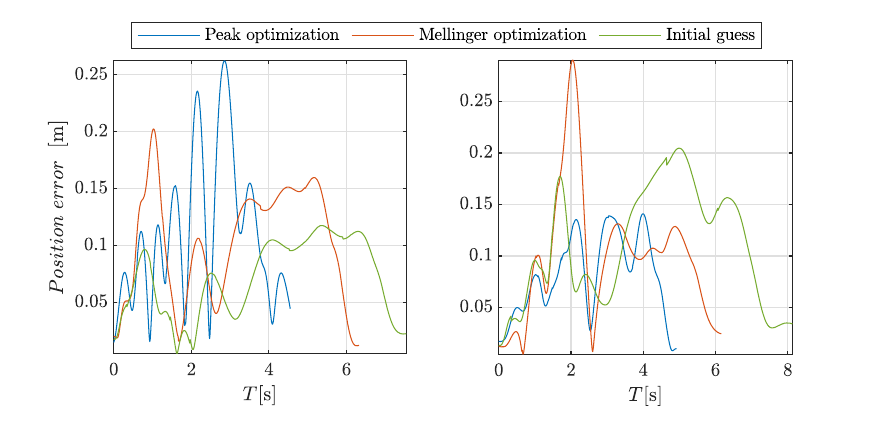}
    \caption{Experimental results, position error of minimum jerk (left) and minimum snap (right) trajectories.}
    \label{fig:Error}
\end{figure}

The peak optimised trajectory is 29\% faster than the Mellinger optimization and 40\% faster than the initial guess. The magnitude of the motion derivatives match between the different methods during the first second, as discussed during the previous section. However it gains a lot of time during the second phase of the trajectory as it is able to fly a lot faster and take a shorter trajectory. The total time had to be longer for the initial guess and Mellingers optimization to achieve feasibility in the first segments. 
The generated trajectories are used as a series of setpoints over time and the drone uses a low-level controller to reach this setpoint. The drone starts to lag behind over time when very aggressive trajectories are used, which results on non-negligible position error. The position error for all trajectories is shown in Fig. ~\ref{fig:Error}. The RMS errors are 10.2 $\mathrm{cm}$, 11.4 $\mathrm{cm}$ and 7.9 $\mathrm{cm}$ for the peak optimization, Mellinger optimization and the initial guess, respectively.

\subsection{Minimum snap trajectories}
The peak optimised trajectory is 20\% faster than the Mellinger optimization and 39\% faster than the initial guess. The differences between the time optimization methods are the same as when the polynomial coefficients were optimised for jerk. However, this time the initial guess deviates significantly from the other methods as it flies a significantly longer trajectory, as it is then able to fly in a very wide arc that minimises the snap, compared to the tighter corners the other methods use. This results in higher speeds compared to the other methods, but does not make up for the additional distance that has to be covered. The RMS errors are 9.0 $\mathrm{cm}$, 11.7 $\mathrm{cm}$ and 12.2 $\mathrm{cm}$ for the peak optimization, Mellinger optimization and the initial guess respectively. The initial guess and the Peak optimized minimum snap trajectories are both slower than their minimum jerk equivalent, however the Mellinger optimized is actually faster than its minimum jerk equivalent.

\begin{table}[h!]
\begin{center}
\caption{Time taken to complete the full trajectory}
\label{tab:finaltime}
\begin{tabular}{  m{16em} | m{1cm}| m{1cm}  } 
Minimised derivative / segment times optimization method & Jerk [s] & Snap [s]\\ \hline
 Initial guess     & 7.54 & 8.13 \\ 
Mellinger optimization      & 6.32 & 6.16 \\ 
Peak optimization  & 4.50 & 4.92 \\
\end{tabular}
\end{center}
\end{table}
\vspace{-0.5cm}

\subsection{Discussions}

It can be concluded from the previous section that the most aggressive trajectories possible within the minimum jerk/snap framework and with the available methods can be generated using the the peak optimization method where the polynomial coefficients are optimised for minimum jerk. The peak optimization does not guarantee feasibility while providing the most aggressive trajectories possible as different limits for acceleration and jerk had to be used for jerk optimized and snap optimized trajectories. It is not possible to give a guarantee of feasibility when flying on the limit due to the thrust being coupled to both the acceleration and the snap. Therefore the maximum thrust could be exceed when it is on the limit in both acceleration and snap. 

However, when it is not required to obtain the most aggressive trajectory possible, the limits could be lowered and give a guaranteed feasible trajectory. This is not possible with the other previously discussed methods. The limits could also be lowered when lower RMS position errors are desired, when manoeuvring through tight spaces for example.

\section{Conclusion}\label{Chapters:Discussion}
A new optimization method for time-optimal trajectory generation under the framework of continuous-time polynomials approach was presented. This new method focuses generating minimal time feasible trajectories. This is achieved by optimising the segment times and polynomials in an alternating where it generates trajectories that are consistently close to the limits of the drone for each motion derivative up to the fourth order, which are directly linked to the control inputs by differential flatness. The new and existing methods were implemented in simulation and proven by real world experiments using the Crazyflie platform. The peak optimization was able to achieve the lowest total running time for the all the compared continuous time polynomial optimization methods. The limits used by the peak optimization method can be raised to achieve more aggressive motion and reduce the running time, or be lowered to achieve guaranteed feasible trajectories.

\bibliographystyle{IEEEtran}
\bibliography{Bibliography}



\end{document}